\documentclass[a4paper,fleqn]{cas-dc}

\usepackage[authoryear]{natbib} 
\usepackage{footmisc}
\usepackage{gensymb}
\usepackage[T1]{fontenc}
\usepackage[utf8]{inputenc}

\def\tsc#1{\csdef{#1}{\textsc{\lowercase{#1}}\xspace}}
\tsc{WGM}
\tsc{QE}
\tsc{EP}
\tsc{PMS}
\tsc{BEC}
\tsc{DE}

\begin{document}

\let\WriteBookmarks\relax
\def\floatpagepagefraction{1}
\def\textpagefraction{.001}

\shorttitle{Automatic White Shrimp Biometrical Analysis Using Computer Vision and Deep Learning}

\shortauthors{A. Remache González et~al.}

\title [mode = title]{IMASHRIMP: Automatic White Shrimp (\textit{Penaeus vannamei}) Biometrical Analysis from Laboratory Images Using Computer Vision and Deep Learning} 


\author[1]{Remache González Abiam}
\ead{abiam.remache101@alu.ulpgc.es}
\fntext[1]{Corresponding author. Tel.: +34 603 68 33 05.}
\fnmark[1]

\author[2]{Chagour Meriem}

\author[2]{Bijan Rüth Timon}

\author[2]{Trapiella Cañedo Raúl}

\author[2]{Martínez Soler Marina}

\author[2]{Lorenzo Felipe Álvaro}

\author[2]{Shin Hyun-Suk}

\author[2]{Zamorano Serrano María-Jesús}

\author[4]{Torres Ricardo}

\author[5]{Castillo Parra Juan-Antonio}

\author[4]{Reyes Abad Eduardo}

\author[3]{Ferrer Ballester Miguel-Ángel}

\author[2]{Afonso López Juan-Manuel}

\author[1]{Hernández Tejera Francisco-Mario}

\author[1]{Penate-Sanchez Adrian}

\affiliation[1]{organization={Institute SIANI, Universidad de Las Palmas de Gran Canaria (ULPGC)},
    city={Las Palmas},
    postcode={35001}, 
    country={Spain}}

\affiliation[2]{organization={Aquaculture Research Group (GIA), Institute of Sustainable Aquaculture and Marine Ecosystems (IU-ECOAQUA), Universidad de Las Palmas de Gran Canaria (ULPGC)},
    city={Telde},
    postcode={35413},
    country={Spain}}

\affiliation[3]{organization={Technological Centre for Innovation in Communications (iDeTIC), Universidad de Las Palmas de Gran Canaria (ULPGC)},
    city={Las Palmas},
    postcode={35017},
    country={Spain}}

\affiliation[4]{organization={PRODUMAR S.A.},
    city={Durán},
    postcode={091650},
    country={Ecuador}}

\affiliation[5]{organization={Biotechnology and Marine Genetic S.A. (BIOGEMAR S.A.)},
    city={San Pablo, Santa Elena},
    postcode={090350},
    country={Ecuador}}
    

\def\errorcmfinal{0.07}
\def\dtcmfinal{0.1}

\def\totalerrormapefinal{3.58}
\def\totalerrorcmfinal{0.51}

\def\povherror{0.97}
\def\povcerror{0}

\def\rosherror{12.46}
\def\roscerror{3.64}

\def\datasettotal{12367}
\def\datasettest{1236}

\begin{abstract}
This paper introduces IMASHRIMP, an adapted system for the automated morphological analysis of white shrimp (\textit{Penaeus vannamei}), aimed at optimizing genetic selection tasks in aquaculture. Existing deep learning and computer vision techniques were modified to address the specific challenges of shrimp morphology analysis from RGBD images. IMASHRIMP incorporates two discrimination modules, based on a modified ResNet-50 architecture, to classify images by the point of view and determine rostrum integrity.  It is proposed a "two-factor authentication (human and IA)" system, it reduces human error in view classification from $\povherror$\% to $\povcerror$\% and in rostrum detection from $\rosherror$\% to $\roscerror$\%. Additionally, a pose estimation module was adapted from VitPose to predict 23 key points on the shrimp's skeleton, with separate networks for lateral and dorsal views. A morphological regression module, using a Support Vector Machine (SVM) model, was integrated to convert pixel measurements to centimeter units. Experimental results show that the system effectively reduces human error, achieving a mean average precision (mAP) of 97.94\% for pose estimation and a pixel-to-centimeter conversion error of $\errorcmfinal$ ± $\dtcmfinal$ cm. IMASHRIMP demonstrates the potential to automate and accelerate shrimp morphological analysis, enhancing the efficiency of genetic selection and contributing to more sustainable aquaculture practices.The code are available at \href{https://github.com/AbiamRemacheGonzalez/ImaShrimp-public}{ImaShrimp}
\end{abstract}


\maketitle

\section{Introduction}

Aquaculture provides a sustainable source of aquatic food that meets the nutritional demands of modern societies. In 2022, for the first time in history, aquaculture production surpassed that of capture fisheries, and it is expected to continue expanding in the coming years. This intensification and expansion of aquaculture require improvements in data collection, as well as analytical tools and methodologies.

At the species level, white-leg shrimp (Penaeus vannamei) led global aquaculture production in 2022, with 6.8 million tonnes produced. Ecuador emerged as the world's leading exporter, in large part due to its sustained efforts to modernize production systems and implement genetic breeding programs \cite{fao2024sofia}. In response to the anticipated increase in global shrimp demand, Ecuadorian shrimp farms must further enhance their competitiveness. A key strategy involves the adoption of non-invasive measurement methods in both production and breeding programs. These methods have been shown to reduce costs, improve measurement efficiency, and improve overall quality of the final product \cite{IMAFISH}.

The PMG-BIOGEMAR© genetic breeding program, developed by the University of Las Palmas de Gran Canaria (Spain), has been implemented by the Almar Group, a major shrimp producer based in Ecuador \cite{Shin2020}. Genetic selection is carried out using the Best Linear Unbiased Prediction (BLUP) methodology, in which thousands of shrimp are assessed for growth and morphological traits. Previous studies have established the genetic parameters of these morphological traits and identified the most relevant ones for selection in this population \cite{SHIN2023101649} \cite{MARTINEZSOLER2024741228}. However, the trait measurements in these studies were obtained manually, a process that is time-consuming and prone to errors that are difficult to correct. Automating these measurements can significantly reduce operational costs and allow the evaluation of a much larger number of individuals, thereby strengthening the effectiveness of the genetic selection process. In this study, we propose a novel solution to automatically measure shrimp using deep learning techniques, enabling precise and robust phenotyping.

In this paper, we provide a solution to automatically measure shrimp using deep learning techniques to produce precise and robust measurements. We take advantage of the successful line of research pioneered by \cite{Wei_2016_CVPR} for human pose inference. Our system is designed to predict the coordinates of 23 key points in shrimp images. These key points collectively construct a virtual skeletal structure that encapsulates the morphological characteristics of interest. By mapping these points, the system effectively represents the anatomical framework of the shrimp, allowing the extraction and analysis of relevant morphological variables. 
In our case, we learn how to estimate a set of key points that are comprised of each of the positions from which the measurements must be taken. For example, to measure the length of the head of a shrimp we estimate the virtual skeleton of the animal first and afterward measure in 2D the distance between the tip and the back of the head, which are both part of the skeleton. Once the 2D key points have been detected and correctly estimated, we perform a regression that transforms that 2D measurement to the required 3D measurement. Our system is able to perform both the required lateral and dorsal measurements.

Our proposed method aims to replace the human operators that perform the measurements. Part of said process by humans comprises a level of human error; humans will introduce a certain percentage of data in the system incorrectly. We give a solution to these errors that strongly mitigates the amount of incorrect data due to human error. 

Our system has an artificial intelligence detection system that can predict whether the shrimp is in a lateral or dorsal position. This works as a secondary system that detects when the human might have produced an error and generates an alert so the error in the input of data into the system can be averted. 

Another way in which our system avoids human error is by detecting the \textit{rostrum} of each shrimp. The \textit{rostrum} is the beak-shaped structure that shrimp have on the top of their head. The rostrum enables the measurement of the cephalotorax lenght, which is one of the most intriguing morphological traits to incorporate into genetic breeding programs for the species \cite{MARTINEZSOLER2024741228}. The problem lies in that many shrimp have that horn broken during their lives. For such a reason, detecting if the \textit{rostrum} is present can reduce human error. We use the same approach; if our AI detector predicts that the human could have committed an error, we raise an alarm. 

In summary, the main contributions of our work are the following:

\begin{itemize}
    \item \textbf{An automatic and robust system that can measure shrimp.} By modelling the task of measuring the animal into a task of key point detection, we can robustly measure in 3D the length, height and width of the shrimps body for both lateral and dorsal point of view images. We use a solution based on deep learning techniques, for which we have annotated more than $12$ thousand images.

    Not only lenght, but also height and widht. Maybe is better use dorsal? To be consistent with the rest of the paper
    \item \textbf{Two AI detection systems capable of mitigating human error.} Our system has two complementary AI detection system that can robustly detect the position of the animal and if the \textit{rostrum} is present and alert the human operator of a possible error in the data input.
    \item \textbf{A regression estimator to increase the precision of the 3D measurements.} We further enhance the precision of our methods by learning a regression that can reduce the error in key point virtual skeleton detection.This approach reduces the mean error of the whole system using regression for all their pixel measures    
\end{itemize}

\begin{figure*}[t]
	\includegraphics[width=0.99\textwidth]{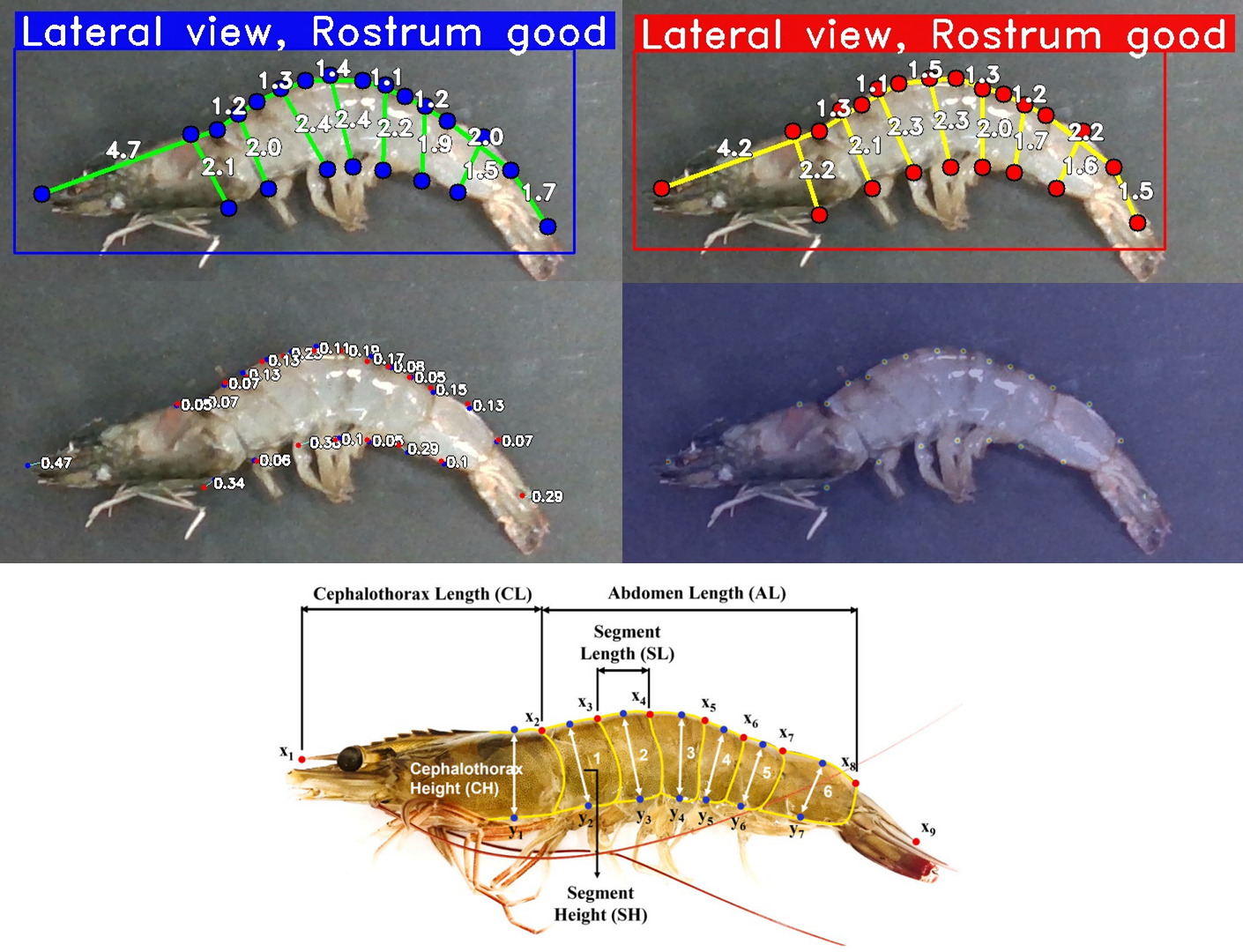} 
 \vspace{-0mm}
    \caption{{\bf Description of the key point virtual skeleton used by our shrimp pose estimator and the measurements performed on the animal used for genetic improvement.} \textbf{Top row:} the image on the left shows the ground truth virtual skeleton used by our method, while the image on the right shows the prediction by our neural network. \textbf{Middle row:} on the left, we can see the error between the ground truth and our estimations. The image on the right shows the heatmaps produced by the neural network for each point. \textbf{Bottom row:} Here we show the measurements that are performed for the lateral view of the shrimp. A full  description of all measurements can be found in \cite{SHIN2023101649} and in \cite{MARTINEZSOLER2024741228}.
    }
\vspace{-0mm}
\label{fig:background_figure}
\centering
\end{figure*}

\section{Related work}

We will now detail the context in which our work resides. We will first detail other works performed in the field of aquaculture engineering that address the extraction of morphological/biometrical information from animals in a wide array of contexts. Secondly, we will describe the work performed in the field of articulated 3D pose estimation, which is the line of work from which our 3D shrimp pose estimation stems. To our knowledge, no one has performed 3D shrimp pose estimation, the closest work was that of \cite{CHIRDCHOO2024200434} that only performs 2D visual analysis and reports errors of $2.1$ cm of mean average error (MAE) for the length of the shrimp, while we obtain 3D estimates and report a MAE of $\totalerrorcmfinal$ cm when calculating the length of a shrimp.

\subsection{Computer vision in morphological analysis}

The morphological measurement of aquatic animals is a critical aspect of fisheries management, species monitoring, and aquaculture. Although advanced artificial intelligence techniques and automated systems have been widely applied to fish species, similar approaches for shrimp remain underexplored.

Few studies have focused specifically on the morphological analysis of shrimp. For example, \cite{shrimp_article_1} and \cite{shrimp_article_5} employed traditional image processing techniques, such as segmentation based on the intensity threshold, to estimate the lengths of shrimp. Other studies, such as \cite{shrimp_article_3} and \cite{shrimp_article_4}, rely on convolutional neural networks to extract a single morphological variable, total length, from the bounding boxes. Specifically, \cite{shrimp_article_3} uses SLCNet \cite{Liu2022SemiSupervisedMI}, and \cite{shrimp_article_4} applies Mask R-CNN \cite{He2017MaskR}. The studies focus solely on pixel-based measurements, lacking a module for physical unit conversion. These findings underscore the potential for automation,n, but highlight significant gaps, particularly in dataset availability and methodological standardization.

In contrast, research on fish species has made substantial progress. Many studies, such as \cite{65e4230f795b6b1f94c9b37c} (which uses a variation based on YOLO \cite{Redmon2016YouOL} called YOLACT \cite{Bolya2019YOLACT}) and \cite{FishAuctionMonitoring} (which applies Mask R-CNN \cite{He2017MaskR}), employ semantic segmentation techniques to estimate a single morphological variable, usually the total length. This simplification is often sufficient when multiple features are not of interest. Other studies, such as \cite{10.1093/icesjms/fsz186} and \cite{9188604}, also use Mask R-CNN \cite{He2017MaskR} for semantic segmentation, followed by total length estimation.

Pose estimation networks have also been applied in morphological studies, as seen in \cite{voskakis2021deep} and \cite{dong2023detection}. \cite{voskakis2021deep} measures two morphological variables, the distance from the mouth to the eye and the distance from the mouth to the fin in two species of fish, using a neural network for estimating poses called Open Pose \cite{Cao2021OpenPose}. Meanwhile, \cite{dong2023detection} estimates seven morphological variables using first YOLO \cite{Redmon2016YouOL} and then for pose estimation,, Lite-HRNet \cite{Yu2021LiteHRNet}. However, the latter study \cite{Yu2021LiteHRNet} focuses solely on the accurate detection of key points without implementing pixel-to-centimetre conversion. To capture multiple morphological features, our study uses pose estimation techniques to detect 23 key points from two different points of view, modifying the VitPose network \cite{xu2022vitposesimplevisiontransformer} for this specific task.

Regarding pixel-to-centimetre conversion modules, there are several methodologies. Some studies, such as \cite{10.1093/icesjms/fsz186} and \cite{dong2023detection}, do not perform physical measurements, instead prioritizing the detection of precise morphological variables. Other works, such as \cite{tonachella2022affordable} and \cite{voskakis2021deep}, employ triangulation systems using a chessboard and binocular cameras for calibration. In addition, some studies use a known real-world measurement to derive a scaling factor, as seen in \cite{FishAuctionMonitoring} and \cite{65e4230f795b6b1f94c9b37c}. Finally, regression models, as utilized in \cite{10.1093/icesjms/fsz186}, offer another approach to efficient conversion from pixel to centimetre. Given the substantial amount of real data collected for the 23 morphological variables, our study opts for a regression model to achieve the conversion.

In these studies, error rates in morphological measurements are commonly evaluated using the Mean Average Percentage Error (MAPE) metric, which provides a scale-independent measure of error. For instance, our shrimp measurements achieve a MAPE of $\totalerrormapefinal$\% for total length, compared to 8\% in \cite{65e4230f795b6b1f94c9b37c}, 3.15\% for sea bream in \cite{voskakis2021deep}, 7.14\% for sea bass in \cite{voskakis2021deep} and 14.57\% for Lito\textit{Penaeus vannamei} shrimp in \cite{CHIRDCHOO2024200434}. These results demonstrate that our method achieves a low error rate in the total length estimation. Moreover, using pose estimation, our system can calculate not only the total length but also 23 morphological variables, offering a comprehensive solution for morphological analysis.

\subsection{Articulated 3D pose estimation}

Obtaining the pose of a human or an animal has been extensively researched for years and great advances have been made. The pose of humans or animals is in essence an articulated skeleton, and the task lies in finding with precision the 3D locations of the joints of said skeletons.

Before deep learning, the most promising results obtained in 2D human pose estimation can be seen in \cite{andriluka2014pck}. Obtaining precise 3D measurements remained a challenge at that time. With the appearance of the Convolutional Pose Machines (CPM) work by \cite{Wei_2016_CVPR} it was showed that using deep learning could yield very robust 3D estimations to the articulated pose estimation problem. This advancement was made possible by creating datasets that addressed the need for larger amounts of training data, some of these datasets were Human 3.6, by \cite{h36m_pami} and HumanEva, by \cite{HumanEva}. From the original CPM research paper, many others continued to improve the 3D estimation. For example, in the work by \cite{Tome_2017_CVPR} it is proposed to optimize the 2D and 3D positions together to improve both tasks through the inherent sharing of information within the neural network architecture. In the work by \cite{MorenoNoguer20163DHP} the 3D pose is obtained by modeling the problem as a regression between two Euclidean distance matrices. In subsequent years, the paradigm changed from using convolutional neurons to using visual by transformers \cite{dosovitskiy2020vit}. This paradigm change was then incorporated into many human pose estimation approaches like \cite{dosovitskiy2020vit} and \cite{xu2022vitposesimplevisiontransformer}.

Concerning the application of said approaches to animal pose estimation, there has been less work overall, but the said techniques have been demonstrated to be robust enough to handle dog poses, like \cite{Rueegg2022BARCLT}, or even zebra, tiger, elephant, and horse, as seen in \cite{yao2022lassie} amongst others. Again, this has been possible due to the creation of datasets like the ones by \cite{Xu_2023_ICCV} and by \cite{PAIRR24Mdataset}. The majority of the articulated 3D animal pose estimation has been focused on mammals; this makes our work quite unique as it shows that such techniques can be used on a wider array of animal species, and particularly those of great economical interest.

\section{Background}

\begin{figure}[t]
	\includegraphics[width=0.99\columnwidth]{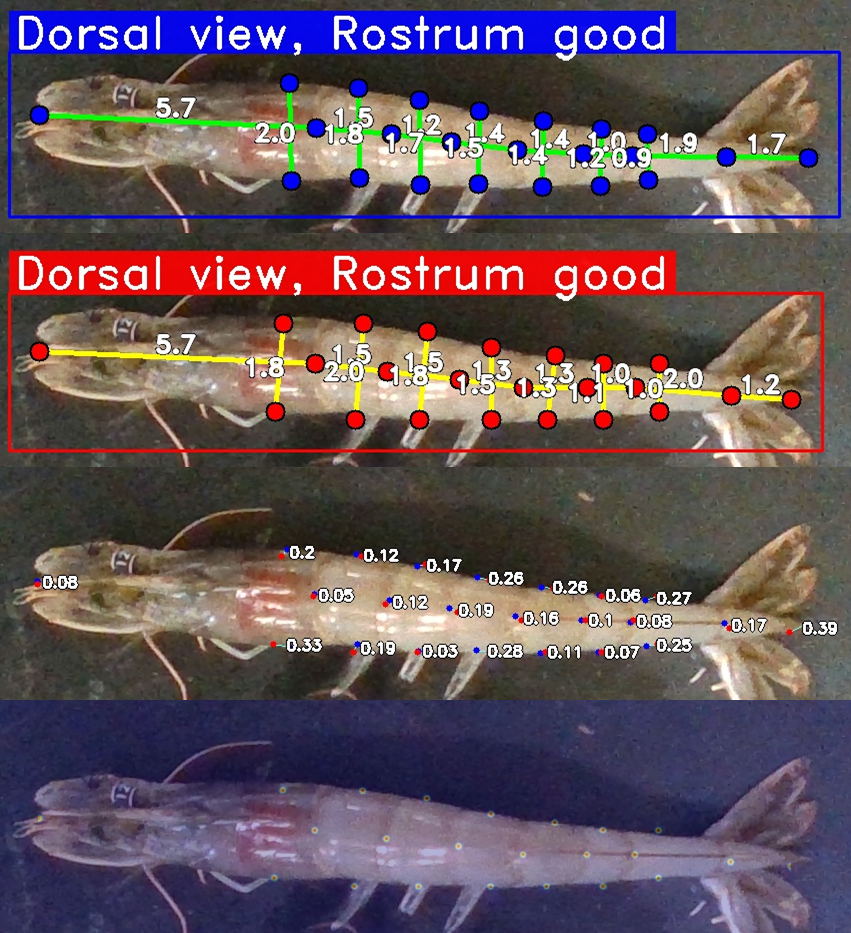} 
 \vspace{-3mm}
    \caption{{\bf Description of the dorsal key point virtual skeleton used by our shrimp pose estimator.} \textbf{First row:} Ground truth virtual skeleton used by our method, with measures in centimetres. \textbf{Second row:} prediction of our neural network pose estimation  with measures in centimetres. \textbf{Third row:} Error between the ground truth and our estimations in centimetres. \textbf{Fourth row:} Heatmaps produced by the neural network for each point.
    }
\vspace{-0mm}
\label{fig:dorsal_skeleton}
\centering
\end{figure}

In this section, we will describe the scenario in which our system works to facilitate understanding of our paper. We will introduce the way in which genetic selection is performed, we will show the image capture setup (in Fig. \ref{fig:capturing_method}) we have used for our data collection and that is used for testing, and finally, we will describe the key point virtual skeleton definition that we propose and that comes directly from the initial and final measurements that are performed by the experts. 

\noindent{\bf Shrimp selective breeding.} Shrimp selective breeding. Genetic selection breeding programs for the species allow breeders to be selected according to their Estimated Breeding Values (EBVs) for a desired trait to obtain the next generation. For the current population, weight and morphological traits are among the most important traits to be selected. To perform such selection, a statistical analysis of morphological traits using BLUP is used. Some of the more costly morphological traits to obtain are the precise measurements of each part of the shrimp. Our system uses 23 morphological measurements from both the dorsal and lateral views of the animal. A detailed description diagram of the morphological measurements, and how they are related to the shrimp virtual skeleton, can be found in Fig.~\ref{fig:ground_truth}. More qualitative examples of the morphological measurements as they are performed in real cases can be seen in Fig.~\ref{fig:background_figure} for the lateral case and in Fig.~\ref{fig:dorsal_skeleton} for the dorsal case.

\noindent{\bf Shrimp virtual skeleton definition.} As explained before, each of the measurements done on the shrimp for selective breeding consists of a starting and final point of measurement. Many of those points are used several times. For example, when measuring the head of the shrimp we measure from point $x_1$ till point $x_2$, and afterwards when measuring the length of the first segment of the body of the shrimp we measure from point $x_2$ to point $x_3$. By taking each of the points required for measurement and their topology we have our proposed virtual skeleton as seen in Fig.~\ref{fig:background_figure} (top row on the left). In our work we estimate two virtual skeletons depending on the point of view, the lateral skeleton, Fig.~\ref{fig:background_figure}, and the dorsal skeleton, Fig.~\ref{fig:dorsal_skeleton}. Given the skeletons that we have defined, we can perform key point pose estimation, similar to the one used in humans in \cite{Wei_2016_CVPR} or \cite{xu2022vitposesimplevisiontransformer}, to learn to predict the points we require for our measurements. We employ to separate neural networks to learn separately the lateral and the dorsal skeleton.

\section{Method}

The entire IMASHRIMP proposal is composed of different elements, on the one hand, the modules based on artificial intelligence such as the discriminator systems (point of view and rostrum integrity) and the shrimp pose estimation system, and on the other hand the morphological regression module.

Firstly, the function of the discriminator modules for the discrimination of images will be explained according to two main factors, the shrimp's point of view and the shrimp integrity of the rostrum. Secondly, the operation of the shrimp pose estimation module will be explained, which will be responsible for detecting 23 key points for each of the views (lateral and dorsal), in cases where the rostrum is broken, it will only detect 22.
Finally, the morphological regression module will be explained, which is responsible for converting the morphological variables resulting from the detection of key points (pose estimation system) from pixels to c
entimetres. As a result, the morphological variables shown in Fig. \ref{fig:ground_truth} will be obtained.

\begin{figure*}[t]
	\includegraphics[width=0.99\textwidth]{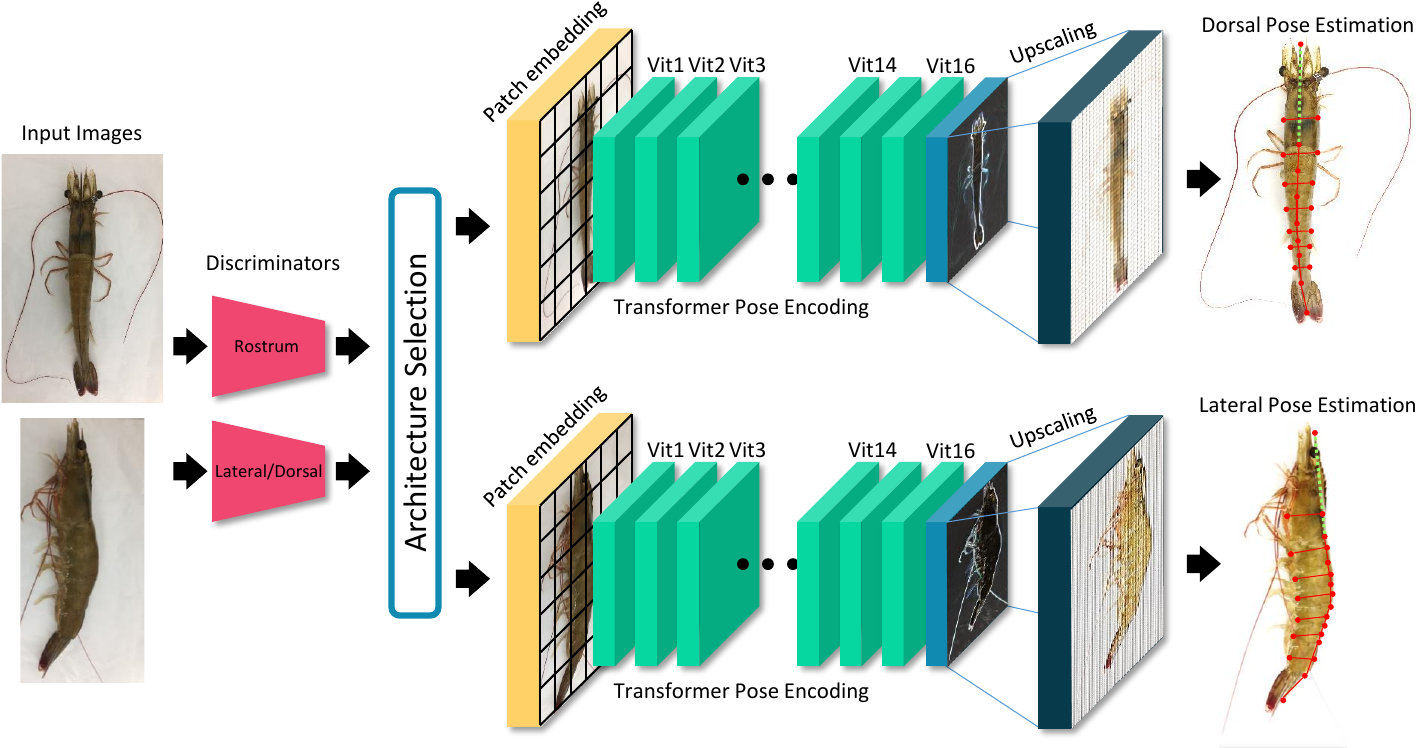} 
 \vspace{-0mm}
    \caption{{\bf Overview of the proposed method for shrimp pose estimation and size regression.} The image shows the architecture of our approach. First, images go through the discriminators to check if the human assessment and the AI assessment is coherent. Based on the view point, two different keypoint virtual skeletons are detected, one for dorsal images and a different one for lateral images. If the rostrum is present or not, those virtual skeletons go from 23 keypoints with rostrum, or 22 keypoints if not. The rostrum part is shown with a green dotted line in the final pose estimation. Concerning the architecture, we can see the patch embeddings of the transformer modules and the upscaling to infers the final positions of the keypoints as described in the method section.
    }
\vspace{-0mm}
\label{fig:proposed_method_figure}
\centering
\end{figure*}

The proposed method consists of the creation of a complete system that integrates the discrimination, pose estimation, and morphological regression modules. The proposed system can be seen in Fig.~\ref{fig:proposed_method_figure}.

\subsection{Discrimination Systems}

As described above, our proposed approach incorporates a set of AI tools that greatly reduce human error. When humans create the metadata associated with the images, they have to introduce if the image was taken from a lateral or dorsal view and if the shrimp has a complete \textit{rostrum}. Based on this information, the image goes through different pose estimation models that return lateral measures or dorsal measures, or reduce the number of key points from 23 to 22 if the \textit{rostrum} is not present.

In order to maximize the robustness of our system, we show that the best approach is to use both the human annotation and the automatically detected AI results. This works as a sort of two-factor authentication, if human and AI agree, the data are introduced into the database, if they disagree and alarm is raised for the data to be checked and corrected. With this scheme, we manage to reduce human error from $\povherror$\% to $\povcerror$\% for the annotation of the view point of the image (lateral/dorsal), and we reduce human error in \textit{rostrum} presence from $\rosherror$\% to $\roscerror$\% with our discriminator systems. The complete results can be seen in Table~\ref{tab:combined_pv_ri_class}.

Both our \textbf{rostrum} and \textbf{lateral/dorsal} classifiers work in the same way to help the human technician. To classify between the lateral/dorsal view and the presence of rostrum, we use a ResNet-50 architecture, by \cite{he2016residual}, for binary classification. We pass the complete image to train the classifier to detect the desired prediction, as seen in Fig.~\ref{fig:capturing_method}. The human makes an assessment $\Psi^{h}_{pose}$ for the pose of the shrimp (lateral or dorsal) and an assessment if the \textit{rostrum} is present $\Psi^{h}_{rostrum}$. In parallel, our discriminator architecture makes a parallel assessment $\Psi^{AI}_{pose}$ and $\Psi^{AI}_{rostrum}$. The final decision on whether to raise an alert due to incorrect pose input $\Psi'_{pose}$, or due to \textit{rostrum} $\Psi'_{rostrum}$, is made as follows:

\begin{align}
    \Psi'_{pose} &= \Psi^{AI}_{pose} \oplus \Psi^{h}_{pose} \\
    \Psi'_{rostrum} &= \Psi^{AI}_{rostrum} \oplus \Psi^{h}_{rostrum}
	\label{eq:discriminator}
\end{align}

Using an XOR operator $\oplus$, the alert is raised only when there is a disagreement between the human and the AI. This does not fix the very few cases in which both commit a mistake, and this in any case is a lesser problem. Human errors in repetitive tasks come from humans losing focus, whereas AI errors occur due to different factors. Due to our hybrid two-factor authentication, the AI can fix the most common human errors and vice versa.

\subsection{Shrimp Pose Estimation}

We have modelled the task of measuring the shrimp morphology as that of a shrimp pose estimation task, where each of the joints of key points will be used to estimate the desired measurements. We draw from the rich state-of-the-art advances in human pose estimation, in which the best performing approaches use a neural network to learn to predict a series of heatmaps from which the key point locations are derived. We follow the work of \cite{xu2022vitposesimplevisiontransformer} for our neural network by creating an encoder / decoder architecture. Encoding is done using visual image transformer neurons (VIT) by \cite{dosovitskiy2020vit}. The decoder consists of a bilinear layer followed by a $ReLU$ activation function and a final pose predictor as described in \cite{Xiao_2018_ECCV}.

Given an input RGB-D image of size $X \in R^{HxWx4}$, where $H$ is a height of $192$ pixels and $W$ is a width of $256$ pixels, we perform an initial encoding in a patch embedding space $F_0$ of smaller resolution. Our embedding reduces the resolution by a factor of $d=16$ and has dimensions of $C=1280$, which creates a patch embedding of size $F_0 \in R^{\frac{H}{d}x\frac{W}{d}x{C}}$ which in our work leaves us with $F_0 \in R^{12x16x1280}$. From this initial encoding $F_0$ we apply several VIT layers which consists of a multi-head self-attention layer $\Theta_i$ and then a multi-layer perceptron $MLP_i$. Layer normalization is applied before every $F_i$ block, we represent it by the $\widehat{*}$ symbol. We use $16$ VIT layers for our encoder. The dimensionality of the embedding $C$ remains unchanged throughout the encoding. The final form of the encoding $F_i$, where $i$ represents the encoding in each intermediate step, is as follows:

\begin{equation}
    F_i = F_{i-1} + \Theta_i(\widehat{F_{i-1}}) + MLP_i\{ F_{i-1} + \Theta_i(\widehat{F_{i-1}}) \}
\end{equation}

Our decoder architecture is a simple combination of a bilinear interpolation and a ReLU activation function, with the final pose predictor. Given the final encoding output of our 16 VIT layers $F_{16} \in R^{\frac{H}{16}x\frac{W}{16}x{C}}$, the decoder creates a setup heatmap per each key point $k$ of the virtual skeleton and upscales by a factor of 4. The number of key points in our configuration can be $N_k=23$ for the general case and $N_k=22$ for animals without \textit{rostrum}. This yields the tensor $F_{heat} \in R^{\frac{H}{4}x\frac{W}{4}x{N_k}}$. Given $F_{heat}$ the predictor, optimizes and $L_2$ loss from training data to learn the prediction of the final position of each key point $k$. We show in Fig.\ref{fig:proposed_method_figure} a visual diagram of the shrimp pose estimation network.

In order to train such a neural network, up to $3.6$ million images can be required, as seen in \cite{h36m_pami}. To avoid such high costs in annotation (our dataset is $\datasettotal$ images), we leverage weights trained on different tasks to bootstrap our training through transfer learning. Our encoder has been pre-trained in other tasks/datasets to have better encodings that will allow the pose estimation to be successful. The datasets on which our network has been pre-trained are the MS-COCO dataset, by \cite{lin2015microsoftcococommonobjects}, the AI challenger dataset, by \cite{AI_challenger} and the MPII Human pose dataset, by \cite{andriluka2014pck}.

\subsection{Shrimp Morphological Regression}

Once the key points have been found in the 2D image space, the next step is to derive the real 3D distances. In order to do so, the simple way to approach it is by calibrating how many centimetres does one pixel equate to, this was the method used by the IMAFISH method, by \cite{IMAFISH_2015}. We consider this approach our baseline method. However, as described by \cite{10.1093/icesjms/fsz186} it yields much more precise measurements to construct a regression model from known instances. Following their discoveries, we construct a regression model per measurement.

In the general case, our pose estimator returns a set of 23 key points $X_i$ and from those 23 points a set of 22 measurements $D_a$. We refer again to Fig.~\ref{fig:ground_truth} for details of each key point and their related measurements. The desired measurements must be in 3D, which we define as $D^{3D}_a$, but the ones we obtain from our shrimp keypoint detection are in 2D within the image, which we define as $D^{2D}_a$. To obtain the 3D measurements from the 2D, we propose to learn a regression model. If one seeks to find a mapping between our 2D measurements $D^{2D}_a$ and the real 3D measurements $D^{3D}_a$ it can be posed as a problem of learning the coefficients of the function $D^{3D}_a = D^{2D}_a * \alpha + \beta$. Due to the inaccuracies of our data, which come from the human measurements, pixel quantization effects and the camera parameters, an exact solution of ${\alpha,\beta}$ does not exist. Due to this reason regression tries to find the closest hyperplane, as close to flat as possible, that models the relation between our 2D and 3D measurements. It does so by performing the optimization described by \cite{Vapnik96}. We use Vapnik's Support Vector Regression model, as it proved to be the best performing approximation. 

We obtain an individual regression per measurement; the regression function is learned from the training samples of ${D^{3D}_a,D^{2D}_a}$ pairs. If we compare our regression approach with the baseline of just performing a simple calibration, we can see in Table \ref{tab:cm_errors} that there is a substantial advantage to perform regression.

\section{Experimental Setup}
This section will present different batches of experiments that were carried out to validate the presented approach. First, the dataset used in the experiments is described. Then, each module, i.e., the Point of View Discriminator, the Rostrum Discriminator, the Shrimp Pose Estimation and the Shrimp Size Regressor will be validated separately, each with a set of experiments aimed at demonstrating the performance of the different modules. Finally, an overall validation will be conducted for the whole
proposed system.
\subsection{Dataset}
The images used in this article correspond to the SABE (Servicio de Análisis para Acuicultura y Biotecnología de Alta Especialización) laboratory and were captured for eight months (August 2023 to April 2024). The images were captured using a depth camera, Intel Realsense Depth Camera D435. The camera was fixed with a tripod in a zenithal position at 30 centimetres of the plane in a controlled laboratory environment featuring a black background and similar uniform lighting conditions.

The dataset comprises 12 pairs of images per animal specimen, with each pair consisting of an RGB image and its corresponding depth map. The RGB images are stored in PNG format (2.8 MB), while the depth maps are encoded as matrices in NumPy format (16.2 MB). This structure provides a dual representation of each sample, combining visual and spatial information.

\begin{figure}[t]
	\includegraphics[width=0.99\columnwidth]{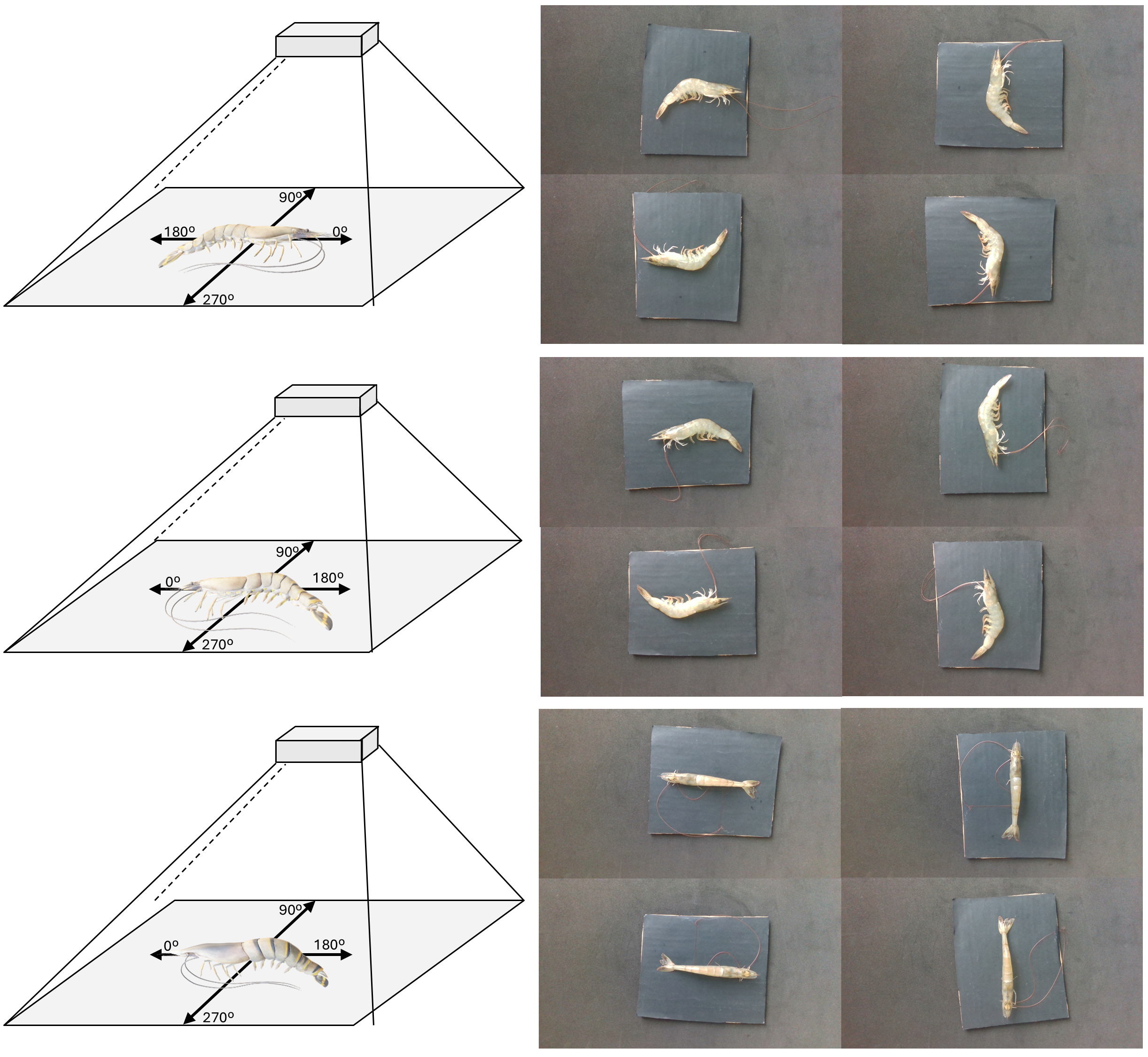} 
 \vspace{-3mm}
    \caption{{\bf Image acquisition setup and methodology.} All shrimp were capturing in four degrees $0\degree$, $90\degree$, $180\degree$ and $270\degree$.\textbf{First row:} Capture of images of the shrimp's right lateral point of view at all degrees. \textbf{Second row:} Capture of images of the shrimp's left lateral point of view at all degrees. \textbf{Third row:} Capture of images of the shrimp's dorsal point of view at all degrees.
    }
\vspace{-0mm}
\label{fig:capturing_method}
\centering
\end{figure}

The same animal is photographed from three different points of view, lateral right, lateral left, and dorsal view at four different angles; see Fig. \ref{fig:capturing_method}.  After the imaging procedure, 2856 images containing rotten shrimp and 2621 images that were blurred were manually discarded. The resulting dataset contains 12367 shrimp images, with a total of 1030 individuals photographed.  

For the discriminator modules, ground truth data were established to ensure a robust verification mechanism. During image capture, researchers concurrently recorded information about the point of view (Lateral or Dorsal) and the rostrum's integrity (Good or Break). This dataset includes annotations from 12367 images, comprising both human observations and those designated as ground truth. The dual-phase documentation approach aims to enhance data reliability by comparing real-time observations with post-acquisition assessments.

\begin{figure}[t]
	\includegraphics[width=0.99\columnwidth]{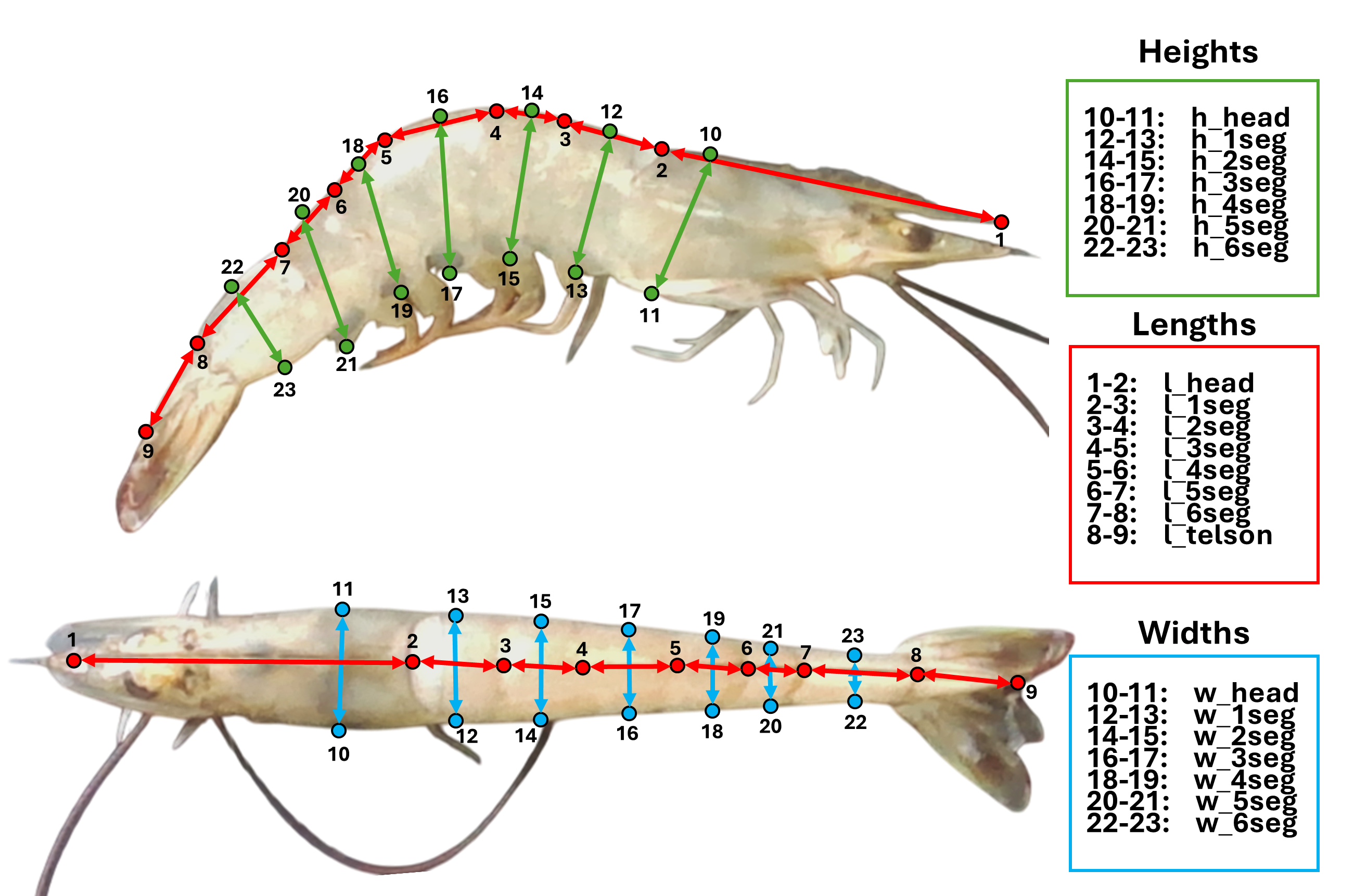} 
 \vspace{-3mm}
    \caption{{\bf Description of the key point virtual skeletons (lateral and dorsal) and the extracted morphological measurements.} \textbf{First row:} Shrimp lateral key point virtual skeleton, key points 1 to 9 can be identified (red points), representing morphological variables of length. key points 10 to 23 can be identified (green points), representing morphological variables of heights. \textbf{Second row:} Shrimp dorsal key point virtual skeleton, key points 1 to 9 can be identified (red points), representing morphological variables of length. key points 10 to 23 can be identified (blue points), representing morphological variables of widths.
    }
\vspace{-0mm}
\label{fig:ground_truth}
\centering
\end{figure}

Of the total number of images captured, 12367 were annotated with 23 key points each for the 'Shrimp Pose Estimation' module. Then, using the key points, we extract the morphological variables of length, height, and width. Examples of ground truth labels and morphological variables can be found in Fig. \ref{fig:ground_truth}. The annotations were created using a generic annotation tool, CVAT (Computer Vision Annotation Tool), and subsequently exported in the COCO key points format \cite{lin2015microsoftcococommonobjects}.

For the Shrimp Size Regressor, during image capture, information was also collected about the actual morphological variables of the animal, identical to those shown in Fig. \ref{fig:ground_truth}. This information will be used in the regression model that will be explained in the following sections.

\subsection{Proposed discriminators experiments}
These study employs a ResNet-50 \cite{he2016residual} inspired neural network to improve the assessment of shrimp morphology through binary discriminators of two different key features: point of view and rostrum integrity. The model's performance will be evaluated by measuring error rates (Error \%) on a test set, comparing errors from: (1) human researchers, (2) the discriminator, and (3) their combined system.

To quantify the improvement offered by the hybrid system, the trained model will be tested specifically on images where human evaluation initially failed. The idea is to demonstrate improvements in accuracy and synergy between human expertise and machine learning.

\subsection{Proposed Pose Estimation Experiments}
The experiments conducted for our Pose Estimation modules aim to evaluate the performance in a test subset and demonstrate its utility for the proposed task. The input images are processed through a dual discriminator system, which assesses the point of view and rostrum integrity. Based on the initial classification, the images are routed to one of four specific processing modules:

\begin{itemize}
\item Lateral Pose Estimation with 23 key points.
\item Dorsal Pose Estimation with 23 key points.
\item Lateral Pose Estimation with 22 key points.
\item Dorsal Pose Estimation with 22 key points.
\end{itemize}

Initially, the four modules are trained independently using type-specific images with corresponding annotation files adapted accordingly. Subsequently, a diverse set of test images, not used during training, will be employed to evaluate the entire system, simulating the normal operation of a fully automated system. We will use evaluation metrics such as Euclidean Point Error (EPE) \cite{epe_article}, Root Mean Square Error (RMSE), Mean Absolute Percentage Error (MAPE) per identified key point of the overall system, i.e. the four modules working as a complete system. For individual modules, we evaluate performance using: (1) Mean Average Precision (mAP) as in \cite{xu2022vitposesimplevisiontransformer}, and (2) Percentage of Correct Key points (PCK) \cite{andriluka2014pck}, which applies a pixel threshold to determine correct key point detection.

\subsection{Proposed size regression experiments}
The Support Vector Machine (SVM) regression model was selected to accurately convert 23 shrimp morphological variables from pixels to centimetres, as illustrated in Fig. \ref{fig:ground_truth}.

To achieve an efficient system for converting pixel measurements to centimetre measurements, two approaches are proposed. First, a non-regression method is implemented using a scale factor derived from ruler images, calculated based on the pixel distance between the 0 cm and 1 cm marks. Second, an SVM regression-based method is developed, leveraging real measurements of all morphological variables in both centimetres (from the image capture phase) and pixels (from annotated ground truth data). The objective is to identify the conversion approach that delivers the most accurate conversion from pixels to centimetres.

As shown in Fig. \ref{fig:ground_truth}, morphological length measurements can be acquired through two standard viewpoints: (1) lateral-view imaging or (2) dorsal-view imaging. Using images in both viewpoints with the same animals, we want to check which view gives the most accurate results and choose them for all the final results.

\section{Results}
This section presents results from both quantitative and qualitative perspectives for all experiments: the Discriminator modules, Pose Estimation module, and Shrimp Size Regressor module. To validate the entire system, we will use a test subset with $\datasettest$ images, approximately 10\% of the total annotated dataset with $\datasettotal$ images.

\subsection{Discriminators results}
\subsubsection{Point of view discriminator results}
Our point of view discriminator was trained to classify shrimp images into lateral and dorsal views using a dataset of $\datasettotal$ annotated images. The network was trained for five epochs with a learning rate of 0.002 and a batch size of 256.

Using the test subset ($\datasettotal$ images), we evaluated error rate, correct and incorrect detections across three classifiers: (1) human experts, (2) artificial intelligence (AI), and (3) our proposed hybrid system. Comparative performance results are presented in Table \ref{tab:combined_pv_ri_class}.

Of the $\datasettotal$ ground truth labels, human researchers made 106 errors (0.85\%). Applying the discriminator to these 106 images reduced errors to 11 (0.08\%). Finally, testing the complete system on the $\datasettest$ images achieved perfect accuracy (1.0), correctly classifying 851 lateral and 385 dorsal images.

\subsubsection{Rostrum integrity discriminator results}
The proposed rostrum discriminator was trained to classify shrimp images based on rostrum integrity (good or broken) using a dataset of $\datasettotal$ annotated images. The network was trained for five epochs with a learning rate of 0.0005 and a batch size of 256.

Using the test subset ($\datasettest$ images), we evaluated the error rate, correctly classified and incorrect detections with three classifiers: (1) human experts, (2) artificial intelligence (AI), and (3) our proposed hybrid system. The comparative results are shown in Table \ref{tab:combined_pv_ri_class}.

Of the $\datasettotal$ real labels, the human researchers made 985 errors (7.96\%). Applying the discriminator to these 985 images reduced the errors to 155 (1.25\%). In the $\datasettest$ test images of the complete system, the humans made 154 failures, the AI 165, but together working as a two-factor authentication system made 45 failures, a percentage error of 4.64\%.

\begin{table*}[H]
\centering
\caption{Classification performance of human, AI, and hybrid approaches for point of view and rostrum integrity discrimination.}
\begin{tabular*}{\textwidth}{@{\extracolsep{\fill}}lccc|lccc}
\toprule
\multicolumn{4}{c|}{Point of view discriminator} & \multicolumn{4}{c}{Rostrum integrity discriminator} \\
\midrule
Classifier & Error (\%) & Correct & Incorrect & Classifier & Error (\%) & Correct & Incorrect \\
\midrule
Human & 0.97 & 1224 & 12 & Human & 12.46 & 1082 & 154 \\
AI & 0.00 & 1236 & 0 & AI & 13.35 & 1071 & 165 \\
\textbf{Human-AI (Ours)} & \textbf{0.00} & \textbf{1236} & \textbf{0} & \textbf{Human-AI (Ours)} & \textbf{3.64} & \textbf{1191} & \textbf{45} \\
\bottomrule
\end{tabular*}
\label{tab:combined_pv_ri_class}
\end{table*}

\subsection{Pose Estimation results}
The pose estimation system includes four image processing modules based on the animal's view (lateral or dorsal) and the rostrum condition (intact or broken). The discriminators classified the $\datasettest$ test images into four groups, assigning each to an appropriate pose estimation module. The performance of each module is summarized in Table \ref{tab:performance_in_test}.

\begin{table}[H] 
\caption{Performance of the pose estimation modules on the test dataset ($\datasettest$ images) depending on the classification of the discriminator modules.}
\label{tab:performance_in_test}
\begin{tabular}{lccc}
\toprule
Model & Nº images & mAP 50:95(\%) & PCK\_10px(\%) \\
\midrule
Dorsal-22 & 49 & 89.85 & 88.22 \\
Lateral-22 & 104 & 97.62 & 86.71 \\
Dorsal-23 & 336 & 89.11 & 85.17 \\
Lateral-23 & 747 & 95.92 & 83.73 \\
\midrule
\textbf{General} & \textbf{1236} & \textbf{93.12} & \textbf{85.96} \\
\bottomrule
\end{tabular}
\end{table}

The pixel predictions from the four modules were compiled into a complete system and compared to the annotated ground truth. Errors for each key point of both views are presented in Table \ref{tab:pixel_errors}.

\begin{table*}[H]
\centering
\caption{Comparative error analysis between real and predicted key points for Lateral and Dorsal point of view}
\begin{tabular*}{\textwidth}{@{\extracolsep{\fill}}lccc|lccc}
\toprule
\multicolumn{4}{c|}{Lateral Point of View} & \multicolumn{4}{c}{Dorsal Point of View} \\
\midrule
Point & EPE(px) & RMSE(px) & MAPE(\%) & Point & EPE(px) & RMSE(px) & MAPE(\%) \\
\midrule
1 & 15.72 ±47.03 & 35.06 & 1.27 & 1 & 13.16 ±21.98 & 18.11 & 1.12 \\
2 & 3.46 ±2.22 & 2.91 & 0.30 & 2 & 4.52 ±2.84 & 3.77 & 0.39 \\
3 & 3.78 ±2.58 & 3.24 & 0.33 & 3 & 4.48 ±2.68 & 3.69 & 0.39 \\
4 & 5.50 ±4.05 & 4.83 & 0.47 & 4 & 5.04 ±3.06 & 4.17 & 0.42 \\
5 & 6.36 ±4.33 & 5.45 & 0.57 & 5 & 6.34 ±3.80 & 5.22 & 0.54 \\
6 & 4.01 ±2.72 & 3.43 & 0.36 & 6 & 4.16 ±2.32 & 3.36 & 0.37 \\
7 & 3.31 ±1.93 & 2.71 & 0.30 & 7 & 3.79 ±1.95 & 3.01 & 0.34 \\
8 & 3.84 ±2.43 & 3.21 & 0.36 & 8 & 3.77 ±1.99 & 3.01 & 0.36 \\
9 & 9.18 ±5.24 & 7.47 & 0.72 & 9 & 9.86 ±5.30 & 7.92 & 0.81 \\
10 & 9.55 ±9.02 & 9.29 & 0.79 & 10 & 12.76 ±11.81 & 12.30 & 1.04 \\
11 & 13.07 ±10.96 & 12.06 & 1.12 & 11 & 13.14 ±12.29 & 12.72 & 1.07 \\
12 & 4.17 ±5.76 & 5.03 & 0.36 & 12 & 5.26 ±4.52 & 4.91 & 0.46 \\
13 & 10.72 ±8.95 & 9.87 & 0.90 & 13 & 5.29 ±4.66 & 4.99 & 0.46 \\
14 & 5.16 ±7.66 & 6.53 & 0.44 & 14 & 4.74 ±2.66 & 3.84 & 0.41 \\
15 & 6.65 ±7.58 & 7.13 & 0.56 & 15 & 4.87 ±3.07 & 4.07 & 0.42 \\
16 & 6.89 ±10.73 & 9.01 & 0.62 & 16 & 5.23 ±3.54 & 4.46 & 0.44 \\
17 & 6.82 ±9.65 & 8.36 & 0.61 & 17 & 5.21 ±3.06 & 4.27 & 0.45 \\
18 & 6.10 ±9.77 & 8.14 & 0.56 & 18 & 4.82 ±2.94 & 3.99 & 0.42 \\
19 & 6.80 ±9.19 & 8.08 & 0.61 & 19 & 5.09 ±2.96 & 4.16 & 0.45 \\
20 & 4.21 ±8.60 & 6.77 & 0.39 & 20 & 4.04 ±2.25 & 3.27 & 0.37 \\
21 & 7.47 ±9.17 & 8.37 & 0.68 & 21 & 4.12 ±2.37 & 3.36 & 0.39 \\
22 & 4.35 ±5.01 & 4.69 & 0.41 & 22 & 6.22 ±3.92 & 5.20 & 0.57 \\
23 & 4.25 ±4.81 & 4.54 & 0.39 & 23 & 6.42 ±4.04 & 5.36 & 0.58 \\
\midrule
\textbf{General} & \textbf{6.53 ±11.96} & \textbf{9.64} & \textbf{0.57} & \textbf{General} & \textbf{6.15 ±7.00} & \textbf{6.59} & \textbf{0.53} \\

\bottomrule
\end{tabular*}
\label{tab:pixel_errors}
\end{table*}

\subsection{Size Regressor Results}
To efficiently convert pixels to centimetres, two methods were evaluated: a reference method using a scale factor based on a known real-world measurement and an SVM regression-based approach. A comparison of these methods is shown in Table \ref{tab:cm_errors}, in which the most interesting metrics are Mean Average Error (MAE) and its standard deviation for each of the morphological variables comparing the real measurements and the estimated measurements.

Morphological lengths were calculated from key points distributed along the shrimp virtual virtual skeleton in both lateral and dorsal views (Fig. \ref{fig:ground_truth}). To determine the optimal view for retrieving length variables, a test subset with the \textbf{same animals} photographed laterally and dorsally was used to compare which view provided the most accurate length measurements. The comparison results are presented in Table \ref{tab:lengths_lateral_vs_dorsal}.

\subsection{End-to-end results}
\begin{figure*}[H]
	\includegraphics[width=0.99\textwidth]{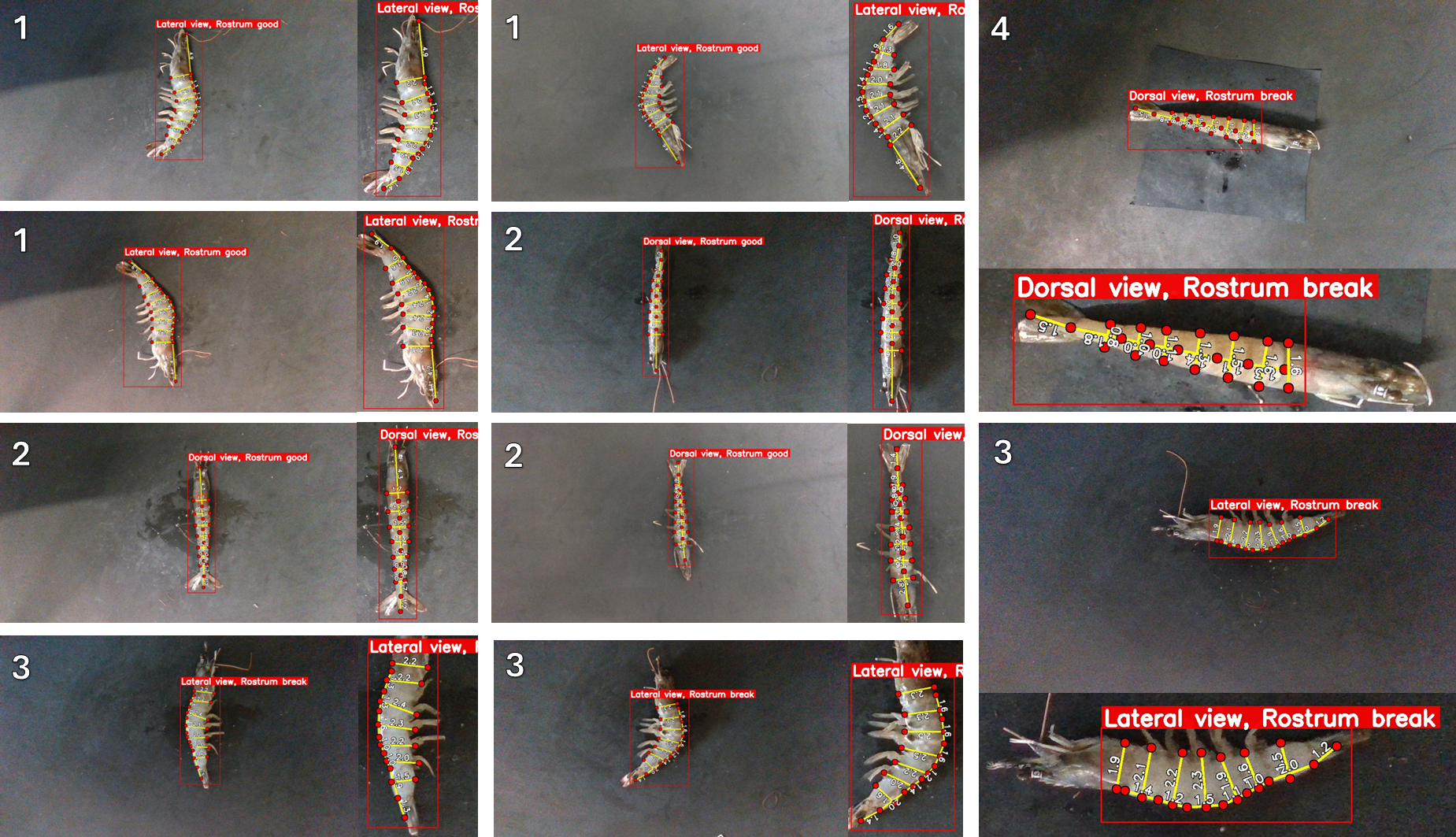} 
 \vspace{-0mm}
    \caption{{\bf Example output images for the proposed system in test dataset}, in which classify information, key points and specimen morphological variables in centimetres are shown for each detected shrimp instance. The images show several results of the test dataset depending on the Pose Estimation Module by which they have been processed: {\bf (1) Lateral Pose Estimation with 23 key points}, {\bf (2) Dorsal Pose Estimation with 23 key points}, {\bf (3) Lateral Pose Estimation with 22 key points} and {\bf (4) Dorsal Pose Estimation with 22 key points}.
    }
\vspace{-0mm}
\label{fig:qualitative_results}
\centering
\end{figure*}

Qualitative results of the complete system are shown in Fig. \ref{fig:qualitative_results}, where key point outputs are combined into a skeleton and visually represented. Furthermore, the size regressor converts pixel-based morphological variables into centimetres.

\begin{figure*}[H]
	\includegraphics[width=0.99\textwidth]{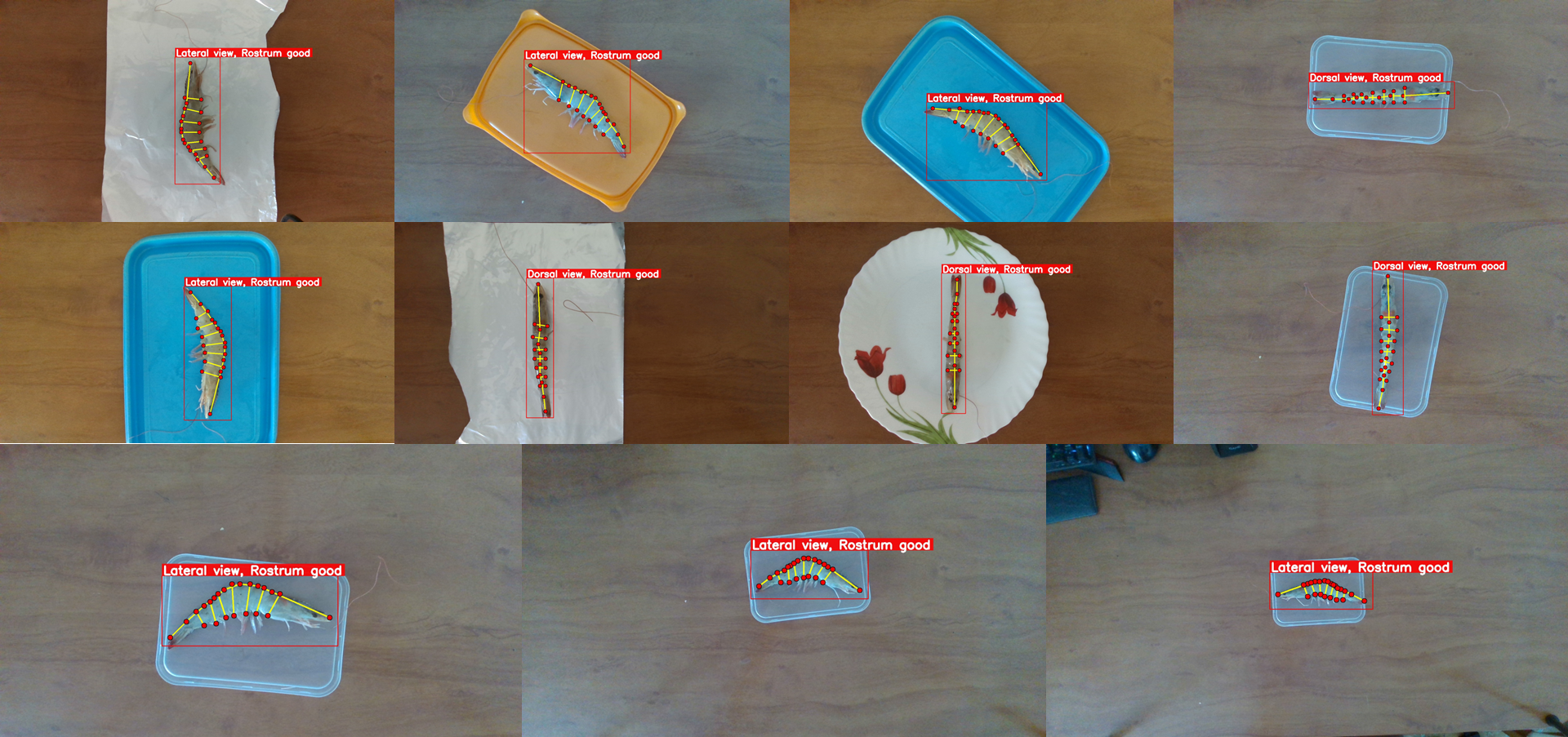} 
 \vspace{-0mm}
    \caption{{\bf Example output images for the proposed system in a different evaluation dataset}, in which classify information and key points are shown for each detected shrimp instance. The images show several results depending on the experiment: {\bf First and second row:} Different backgrounds at the same distance and {\bf Third row:} Different distances to the plane 30 cm, 40 cm and 60 cm.
    }
\vspace{-0mm}
\label{fig:qualitative_results_extra}
\centering
\end{figure*}

To demonstrate the robustness and versatility of the pose estimation system, predictions were performed on a dataset with scenarios not included in training. These include images varying backgrounds and different camera distances. Examples of these results are presented in Fig. \ref{fig:qualitative_results_extra}.

\begin{table*}[t]
\centering
\caption{Comparative analysis of the error between ground truth and IMASHRIMP predicted measurements for all white shrimp morphological variables according to the conversion method.}
\begin{tabular*}{\textwidth}{@{\extracolsep{\fill}}lccc|lccc}
\toprule
\multicolumn{4}{c|}{Conversion not using regression} & \multicolumn{4}{c}{Conversion using regression} \\
\midrule
Variable & MAE(cm) & RMSE(cm) & MAPE(\%) & Variable & MAE(cm) & RMSE(cm) & MAPE(\%) \\
\midrule
total & 1.43 ±1.52 & 2.01 & 10.58 & total & 0.51 ±0.76 & 0.83 & 3.58 \\
abdomen & 0.3 ±0.36 & 0.38 & 3.25 & abdomen & 0.23 ±0.3 & 0.3 & 2.5 \\
\midrule
l\_1seg & 0.09 ±0.11 & 0.12 & 6.94 & l\_1seg & 0.09 ±0.13 & 0.13 & 6.3 \\
l\_2seg & 0.09 ±0.09 & 0.11 & 8.03 & l\_2seg & 0.06 ±0.08 & 0.08 & 5.03 \\
l\_3seg & 0.09 ±0.1 & 0.11 & 6.79 & l\_3seg & 0.07 ±0.09 & 0.09 & 5.14 \\
l\_4seg & 0.17 ±0.1 & 0.19 & 17.21 & l\_4seg & 0.08 ±0.09 & 0.1 & 8.21 \\
l\_5seg & 0.08 ±0.08 & 0.1 & 9.3 & l\_5seg & 0.06 ±0.06 & 0.07 & 6.18 \\
l\_6seg & 0.12 ±0.11 & 0.14 & 6.13 & l\_6seg & 0.05 ±0.07 & 0.07 & 2.83 \\
l\_head & 0.51 ±0.49 & 0.62 & 10.89 & l\_head & 0.16 ±0.2 & 0.22 & 3.42 \\
\midrule
Lengths & 0.19 ±0.3 & 0.3 & 9.43 & Lengths & 0.08 ±0.12 & 0.12 & 5.25 \\
\midrule
h\_head & 0.13 ±0.16 & 0.18 & 6.3 & h\_head & 0.11 ±0.15 & 0.15 & 5.12 \\
h\_1seg & 0.09 ±0.12 & 0.12 & 4.09 & h\_1seg & 0.08 ±0.11 & 0.11 & 3.72 \\
h\_2seg & 0.09 ±0.1 & 0.12 & 4.21 & h\_2seg & 0.06 ±0.08 & 0.08 & 2.76 \\
h\_3seg & 0.1 ±0.1 & 0.12 & 4.39 & h\_3seg & 0.07 ±0.09 & 0.09 & 2.93 \\
h\_4seg & 0.09 ±0.09 & 0.11 & 4.48 & h\_4seg & 0.06 ±0.08 & 0.08 & 2.96 \\
h\_5seg & 0.11 ±0.1 & 0.13 & 5.79 & h\_5seg & 0.07 ±0.09 & 0.09 & 3.73 \\
h\_6seg & 0.04 ±0.06 & 0.06 & 2.95 & h\_6seg & 0.04 ±0.05 & 0.05 & 2.54 \\
\midrule
Heights & 0.09 ±0.11 & 0.12 & 4.6 & Heights & 0.07 ±0.1 & 0.1 & 3.39 \\
\midrule
w\_head & 0.14 ±0.09 & 0.16 & 8.4 & w\_head & 0.07 ±0.08 & 0.08 & 4.18 \\
w\_1seg & 0.15 ±0.11 & 0.18 & 10.26 & w\_1seg & 0.07 ±0.08 & 0.09 & 5.32 \\
w\_2seg & 0.2 ±0.08 & 0.22 & 14.17 & w\_2seg & 0.06 ±0.07 & 0.08 & 4.24 \\
w\_3seg & 0.15 ±0.09 & 0.17 & 12.46 & w\_3seg & 0.06 ±0.07 & 0.07 & 4.45 \\
w\_4seg & 0.12 ±0.08 & 0.14 & 11.38 & w\_4seg & 0.05 ±0.06 & 0.06 & 4.5 \\
w\_5seg & 0.1 ±0.09 & 0.13 & 10.72 & w\_5seg & 0.06 ±0.07 & 0.07 & 6.16 \\
w\_6seg & 0.08 ±0.08 & 0.1 & 11.18 & w\_6seg & 0.05 ±0.06 & 0.06 & 7.22 \\
\midrule
Widths & 0.13 ±0.1 & 0.16 & 11.14 & Widths & 0.06 ±0.07 & 0.08 & 5.35 \\
\midrule
General & 0.13 ±0.2 & 0.2 & 8.2 & \textbf{General} & \textbf{0.07 ±0.1} & \textbf{0.1} & \textbf{4.64} \\
\bottomrule
\end{tabular*}
\label{tab:cm_errors}
\end{table*}

\begin{table*}[t]
\centering
\caption{Comparative error analysis of using Lateral or Dorsal point of view to measure lengths variables}
\begin{tabular*}{\textwidth}{@{\extracolsep{\fill}}lccc|lccc}
\toprule
\multicolumn{4}{c|}{Lateral Point of View Network} & \multicolumn{4}{c}{Dorsal Point of View Network} \\
\midrule
Variable & MAE(cm) & RMSE(cm) & MAPE(\%) & Variable & MAE(cm) & RMSE(cm) & MAPE(\%) \\
\midrule
total & 0.499 ±0.806 & 0.834 & 3.582 & total & 0.357 ±0.451 & 0.500 & 2.544 \\
abdomen & 0.261 ±0.324 & 0.325 & 2.782 & abdomen & 0.285 ±0.333 & 0.343 & 3.019 \\
l\_1seg & 0.089 ±0.115 & 0.118 & 6.440 & l\_1seg & 0.087 ±0.104 & 0.108 & 6.311 \\
l\_2seg & 0.059 ±0.081 & 0.082 & 5.179 & l\_2seg & 0.061 ±0.083 & 0.084 & 5.362 \\
l\_3seg & 0.069 ±0.091 & 0.091 & 5.274 & l\_3seg & 0.062 ±0.089 & 0.089 & 4.790 \\
l\_4seg & 0.068 ±0.080 & 0.085 & 6.978 & l\_4seg & 0.071 ±0.083 & 0.088 & 7.350 \\
l\_5seg & 0.054 ±0.065 & 0.067 & 6.147 & l\_5seg & 0.060 ±0.073 & 0.074 & 6.835 \\
l\_6seg & 0.058 ±0.074 & 0.074 & 3.006 & l\_6seg & 0.069 ±0.089 & 0.089 & 3.580 \\
l\_head & 0.201 ±0.244 & 0.254 & 4.197 & l\_head & 0.186 ±0.243 & 0.244 & 3.830 \\
\midrule
Lengths & 0.091 ±0.133 & 0.134 & 5.270 & \textbf{Lengths} & \textbf{0.088 ±0.129} & \textbf{0.129} & \textbf{5.433} \\
\bottomrule
\end{tabular*}
\label{tab:lengths_lateral_vs_dorsal}
\end{table*}

\section{Conclusion}
The primary contribution of this work is the introduction of IMASHRIMP, a system comprising multiple integrated modules designed to automate and minimize human error in the morphological analysis of white shrimp (\textit{Penaeus vannamei}). To our knowledge, no previous studies have applied pose estimation techniques to shrimp or performed regression analysis of 23 morphological variables using RGB-D images. A key feature is its dual classification modules, which identify shrimp point of view (lateral or dorsal) and rostrum integrity (intact or broken), streamlining the workflow. The results are promising and could improve genetic selection by automating phenotypic analyses, enabling larger population studies with fewer errors.

For the complete test subset of $\datasettest$ images, discriminators operating as a two-factor authentication system reduced human error in point of view discrimination from 0.97\% to 0\% and in rostrum integrity discrimination from 12.46\% to 3.64\%.

The pose estimation module demonstrated robust performance, achieving a mean average precision (mAP) of 93.12\% for the test set. The Euclidean Point Error (EPE) for the 23 key points was 6.53 ±11.96 pixels for the lateral view and 6.15 ±7.00 pixels for the dorsal view.

The pixel-to-centimetre conversion module successfully retrieved morphological variables using two approaches: (1) regression-based analysis and (2) reference-based scaling. The regression-based method outperformed the reference-based scaling method, with a mean average error (MAE) of $\errorcmfinal$ ± $\dtcmfinal$ cm compared to 0.13 ± 0.19 cm, respectively. Furthermore, both lateral and dorsal viewpoints proved to be almost equally accurate in measuring length-related variables, being practically equal with an error of 0.09 cm,

One line of work for the medium-term future would be to use a data set in industrial environments other than the laboratory. In addition, work could be done on a solution that measures with the same precision several shrimps per image. Finally, in the long term, the data obtained could be used in more demanding projects such as 3D pose and shape detection.




\bibliographystyle{cas-model2-names}

\clearpage
\bibliography{references}

\end{document}